\documentclass[final]{cvpr}

\usepackage{times}
\usepackage{epsfig}
\usepackage{graphicx}
\usepackage{amsmath}
\usepackage{amssymb}
\usepackage{verbatim}
\graphicspath{ {figures/} }
\usepackage{changepage}
\usepackage{arydshln}

\usepackage{subcaption}
\DeclareMathAlphabet{\mymathbb}{U}{BOONDOX-ds}{m}{n}
\usepackage{arydshln}
\usepackage[ruled,longend]{algorithm2e}
\usepackage{pifont}
\usepackage{dsfont}
\usepackage{xcolor}
\newcommand{\cmark}{\ding{51}}%

\usepackage{multirow}
\usepackage[pagebackref=true,breaklinks=true,colorlinks,bookmarks=false]{hyperref}

\def\cvprPaperID{51} 
\def\confYear{CVPR 2021}

\newcommand{\myparagraph}[1]{\vspace{2pt}\noindent{\bf #1}}

\DeclareMathOperator*{\argmax}{arg\,max}

\begin{document}

\title{\vspace{-15pt} A Closer Look at Self-training for Zero-Label Semantic Segmentation}

\author{Giuseppe Pastore$^1$, Fabio Cermelli$^{1,2}$, Yongqin Xian$^{3}$,\\ Massimiliano Mancini$^{4}$, Zeynep Akata$^{3,4,5}$, Barbara Caputo$^{1,2}$\\
$^1$Politecnico di Torino, $^2$Italian Institute of Technology, 
$^3$MPI for Informatics, \\ $^4$University of Tübingen, $^5$MPI for Intelligent Systems
}

\maketitle

\begin{abstract}

 Being able to segment unseen classes not observed during training is an important technical challenge in deep learning, because of its potential to reduce the expensive annotation required for semantic segmentation. Prior zero-label semantic segmentation works approach this task by learning visual-semantic embeddings or generative models. However, they are prone to overfitting on the seen classes because there is no training signal for them.  In this paper, we study the challenging generalized zero-label semantic segmentation task where the model has to segment both seen and unseen classes at test time. We assume that pixels of unseen classes could be present in the training images but without being annotated. Our idea is to capture the latent information on unseen classes by supervising the model with self-produced pseudo-labels for unlabeled pixels. We propose a consistency regularizer to filter out noisy pseudo-labels by taking the intersections of the pseudo-labels generated from different augmentations of the same image. Our framework generates pseudo-labels and then retrain the model with human-annotated and pseudo-labelled data. This procedure is repeated for several iterations. As a result, our approach achieves the new state-of-the-art on PascalVOC12 and COCO-stuff datasets in the challenging generalized zero-label semantic segmentation setting, surpassing other existing methods addressing this task with more complex strategies. Code can be found at \url{https://github.com/giuseppepastore10/STRICT}.
 
\end{abstract}

\section{Introduction}





Tremendous progress has been made in semantic segmentation by deep learning~\cite{FCN1,deeplab3,zhao2017pyramid} on large human-annotated datasets~\cite{cocostuff,cordts2016cityscapes}. 
As this requires expensive pixel-wise annotations~\cite{cordts2016cityscapes}, 
reducing the pixel-level supervision becomes important, e.g. weakly supervised~\cite{whatsthepoint,simpledoesit,scribblesup,saliency,ficklenet,panet,oneShot} and few-shot~\cite{panet,oneShot} learning. In the extreme case, the task is zero-label semantic segmentation~\cite{spnet,zs3, cagnet} and the goal is to segment the novel classes not annotated during training. One major limitation of zero-label semantic segmentation is that the model is only evaluated on unseen classes. This is not realistic as any class could be present at test time. In generalized zero-label semantic segmentation~(GZLSS) the model is required to segment both seen and unseen classes. GZLSS is challenging because it suffers from the severe class-imbalanced issue, leading to a significant performance drop on unseen classes. SPNet~\cite{spnet} fixes this issue by reducing the prediction scores of seen classes by a constant factor $\gamma$ that is sensitive and hard to tune. CaGNet~\cite{cagnet} and ZS3~\cite{zs3} propose to learn generative models that synthesize features of unseen classes. Nevertheless, the generated features may have domain shift issues because the true distribution is never observed. It is also worth noting that the training set includes many unlabeled pixels from unseen classes due to the large amount of class co-occurrences. While those unlabeled pixels contain complementary information about unseen classes, SPNet~\cite{spnet} simply ignores them during training.  

\begin{figure}[t]
\centering
    \includegraphics[width=1\linewidth]{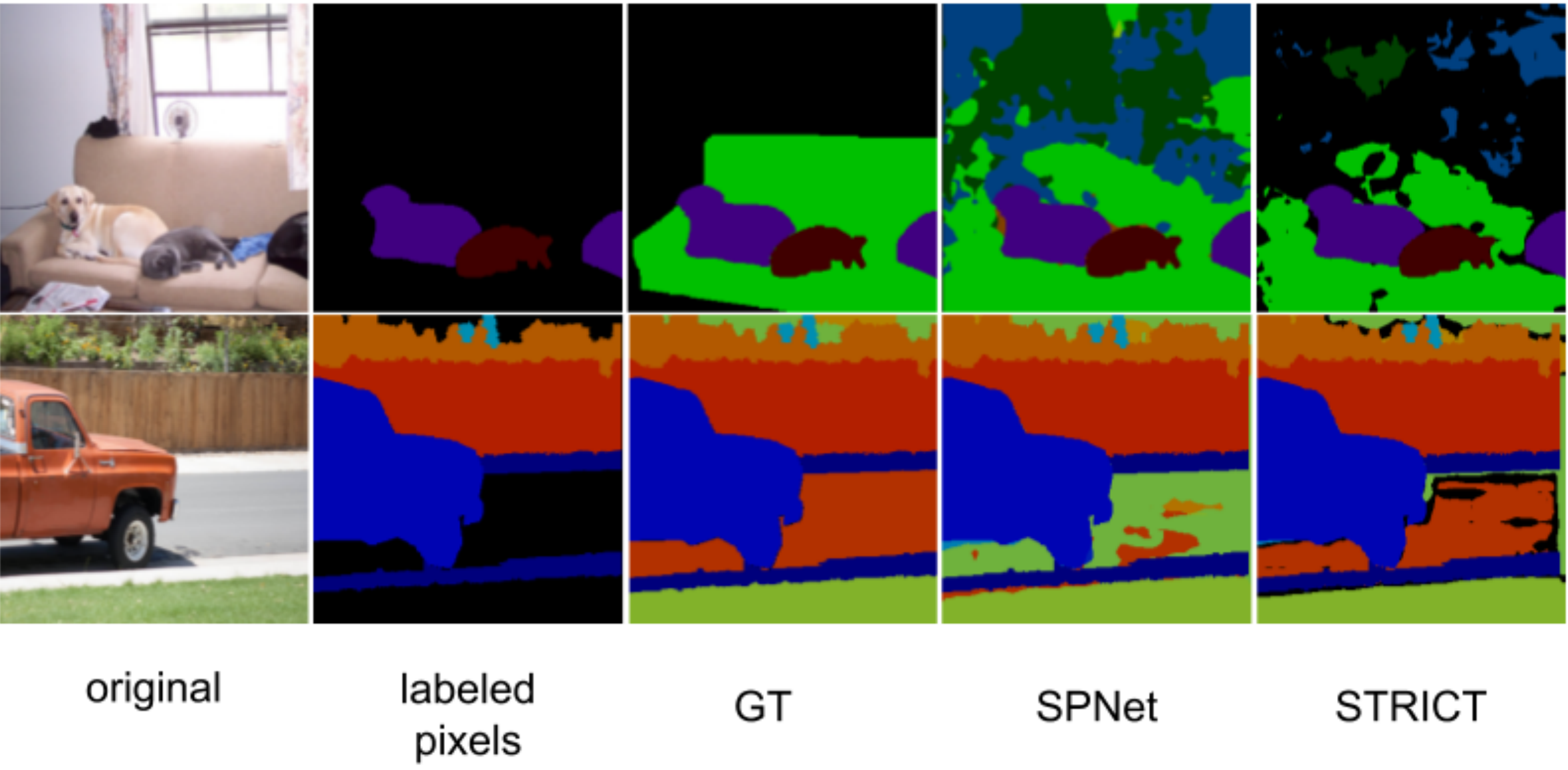}
\caption{In generalized zero-label semantic segmentation, 
some unseen class pixels are ignored although they might be relevant at test-time. 
We approximate the ground truth by pseudo-labeling the unlabelled pixels, i.e. our Self-Training with Consistency Constraint (STRICT) is effective.
In the Figure, \textit{labeled pixels} and \textit{GT} refers respectively to the masked and actual ground truth, \textit{SPNet} and \textit{STRICT} to the pseudo-labeled masks produced by SPNet and our method.} 
\label{fig:teaser}
\end{figure}

We propose to capture the latent information about unseen classes by supervising the model with self-produced pseudo-labels for the unlabeled pixels. 
Pseudo-labeling is not a novel concept, and it has been used as a self-training strategy for semi-supervised learning \cite{pseudoSimpleEfficient,pseudoAndConfrimationBias,metaPseudoLabels,learnToSelfTrain,labelPropagation}.
However, generating accurate pseudo-labels for unseen classes in semantic segmentation is difficult because the accuracy on unseen classes is often much lower than in the supervised case. Consequently,  pseudo-labels generated from a pretrained GZLSS model~(e.g., SPNet~\cite{spnet}) become noisy and may hurt the performance~(as shown in Figure~\ref{fig:teaser}). 
To this end, we introduce an efficient consistency constraint to reduce the noise of pseudo-labels. 
The key assumption is that a pseudo-label is more likely to be correct if the model predicts the same label when presented different augmented versions of the same image. 
While being agnostic to complex hyperparameters tuning and 
robust in pseudo-labeling, our approach provides an effective fine-tuning of the model and a higher predictive capability on the unseen classes. 
By periodically updating the pseudo-label generator with the one fine-tuned through this process, we progressively improve the signal robustness for the unseen classes. 

Our main contributions are: (a) we devise a self-training pipeline to obtain strong supervision for unseen classes from unlabelled pixels in GZLSS. The key component is a pseudo-label generator that enforces consistency constraints on data augmentations ; (b) we show that a model finetuned through such process progressively enhances its ability in predicting unseen classes, and consequently the quality of pseudo-labels.
(c) we extensively analyze our approach on PascalVOC12, both excluding and including the background, and COCO-stuff datasets, demonstrating that our model stands as the state-of-the-art in GZLSS.

\section{Related Works}

\myparagraph{Zero-Shot Learning.}
The ZSL models can be divided in four main categories \cite{zsl_overview} according to how they transfer knowledge from seen to unseen categories.
The first provides a two-stage approach to obtain posterior class probabilities from intermediate attributes extracted from images through additional classifiers \cite{attribute_based}.
The  second  tackles the  task as  a visual-semantic embedding problem evaluating the compatibility between the visual space and the semantic one, so that proximity translates in a semantic relationship \cite{latent_embedding,label_embedding,semOutCode,structuredEmbeddings,semOutCode,similarityFunctions}.
The third category uses a class-level semantic conditioned generator to feed additional synthetic CNN features for unseen classes during the training of a discriminative classifier \cite{featureGenerating,genVisualRepresentations}.
The last category's models address the task in a completely generative way, by modeling the class-conditional distributions to capture semantic relationships among seen and unseen classes \cite{bayesianZSL,expFamily,latentPrototypeModel,ZSRecognition}.
In this work, we refer to the second and third ones as they inspired the existing methods for GZLSS. 

\myparagraph{Generalized Zero Label Semantic Segmentation.} To the best of our knowledge, only SPNet \cite{spnet}, ZS3 \cite{zs3}, and CaGNet \cite{cagnet} directly address GZLSS. SPNet follows the second category's approach \cite{semOutCode}:
a segmentation model is entrusted to extract the visual features that are then projected in the semantic space by a matrix multiplication with a word embedding representation; ZS3 and CaGNet extend the features-generative approach used by \cite{genVisualRepresentations} in classification: the former uses a Graph Convolutional Network to embed a contextual prior on categories disposition (\textit{"mouse is commonly close to the keyboard"}, ...); the latter  does the same but at pixel-wise level, feeding the generator with a contextual latent code instead of the random noise.
ZS3 and CagNet don't directly address the GZLSS scenario, but they propose a variant of their models to do it through self-training, respectively indicated as \textit{ZS5} and \textit{CaGNet + ST}. In this work, we rely on SPNet as we want to demonstrate how an approach as simple as ours can enhance the prediction capability of a segmentation model.

\myparagraph{Self-training in semantic segmentation.}
Pseudo-labeling has been widely used as a self-supervision strategy in poorly annotated computer vision scenarios \cite{pseudoSimpleEfficient,pseudoAndConfrimationBias,metaPseudoLabels,learnToSelfTrain,labelPropagation}.
In image classification, often a pre-trained reference model generates pseudo-labels for unlabelled pixels by embedding a fixed target distribution $q*$ during training \cite{pseudoSimpleEfficient,pseudoAndConfrimationBias} or one continuously adapting to student $p_\theta$'s learning state \cite{metaPseudoLabels}. \cite{labelPropagation} infers pseudo-labels in a transductive setting through label propagation on a nearest neighbor graph built with the features extracted by the model for labeled and unlabeled data.
In semi-supervised semantic segmentation, many works rely on \textit{consistency training}: PseudoSeg \cite{pseudoSeg} generates pseudo-labels for unlabelled pixels by wisely fusing different sources of predictions, decoder and Grad-CAM, and then it imposes the consistency of the predictions of multiple augmented images with such pseudo-labels.
\cite{highLowConsistency} proposes to use adversarial training of a segmentation model that figures as a generator to strengthen the predictions for unlabeled data and to use the discriminator both to identify as good/fake predictions and as a quality measure to select most confident predictions.
\cite{crossConsistency} forces an invariance of the predictions over different encoder's outputs perturbations.
\cite{naiveStudent} shows that iteratively applying pseudo-labeling enhances Scene Segmentation in Urban Video Sequences.
In ZSL, \cite{effetiveDeepEmbedding} subordinates the pseudo-labels selection to the model's confidence, in a transductive ZSL scenario.
In GZLSS, \cite{unbiasedZS,pseudoAndConfrimationBias} enhances the quality of hard pseudo-labeling by first training its model according to an unbiased loss in a transductive way. 
Instead, ZS5 and CaGNet filter out the $p$\% less confident labels self-produced in a GZS setting for the unlabelled pixels. 
We aim to self produce unseen labels for the unlabelled pixels as well, but without introducing any sensitive hypeparameters to improve robustness and generating ZS pseudo-labels. 
\section{Self-training with Consistency Constraints }


GZLSS is particularly challenging because of the severe class-imbalanced issue, leading to a significant performance drop on unseen classes. SPNet~\cite{spnet} fixes this issue by reducing the prediction scores of seen classes by a constant factor $\gamma$. However, $\gamma$ is hard to tune due to its sensibility i.e., a small perturbation of $\gamma$ may lead to a significant change in performance. We argue that the unlabeled pixels ignored during training contain useful information about the unseen classes and incorporating those pixels into training would alleviate the class-imbalanced issue. Thus, we propose a self-training framework that leverages 
those unlabeled pixels by generating pseudo-labels for them.


\begin{figure*}[t]
\begin{center}
\includegraphics[width=1\linewidth]{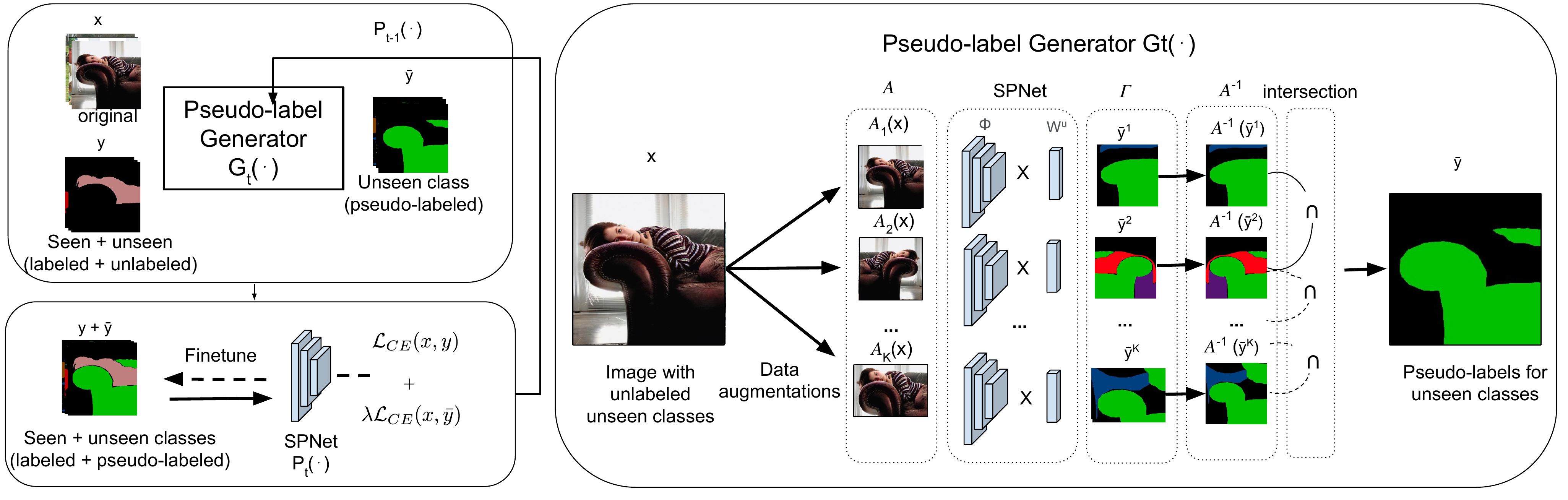}
\end{center}
   \caption{An overview of our STRICT model: 
   at the iteration $t$, the pseudo-label generator $G_t$ produces a pseudo labelled mask $\bar{y}^k$ for the unlabelled pixels of each of the $K$ image' augmentations $\{A_1(x),\ldots,A_k(x)\}$. We then obtain the final pseudo-label mask $\bar{y}$ by applying the intersection operation on them. The model $P_t$ is fine-tuned with the pixel-wise cross-entropy loss computed both on labeled ($y$) and pseudo-labeled ($\bar{y}$) pixels. At the iteration $(t+1)$, $P_t$ will be used for the pseudo-label generator.} 
\label{fig:pipeline}
\end{figure*}

\subsection{Background: semantic projection network}
First, we formally define the task and then describe the semantic projection network~(SPNet)~\cite{spnet}.

\myparagraph{GZLSS Task formulation.} Let $\mathcal{S}=\{1, \ldots, C^s \}$ and $\mathcal{U}=\{C^s+1, \ldots, C^s+C^u\}$ denote two disjoint label sets of seen and unseen classes respectively.  
$\mathcal{T} = \{(x, y)|x \in \mathcal{X}, y_{mn} \in \{\{0\}, \mathcal{S}\}\}$ is the training set where $x$ is an image of spatial size $N\times M$ in the RGB image space $\mathcal{X}$,  $y$ is its corresponding label mask with the same size, and $y_{mn}$ is its corresponding class label at pixel $(m, n)$ belonging to one of the seen classes $S$ or the unlabeled unseen class denoted as $0$. Moreover, each class label is represented by the word embedding~(e.g., word2vec~\cite{word2vec}) associated to its class name. We denote the word embedding matrices of seen and unseen classes with $W^s \in \mathbb{R}^{D\times|C^{s}|}$ and $W^u \in \mathbb{R}^{D\times|C^{u}|}$ with $D$ being the dimension of the word embedding space.
Given $\mathcal{T}$, $W^s$ and $W^u$, the task of generalized zero-label semantic segmentation~(GZLSS) is to learn a model that is capable to make pixel-wise predictions among both seen and unseen classes at test time.

\myparagraph{Semantic projection network.} SPNet~\cite{spnet} consists of a visual-semantic embedding module and a semantic projection layer. The former~(denoted as $\phi$) is based on a CNN backbone~(e.g., DeepLab~\cite{deeplab1}), mapping an input image $x$ to $D$ feature maps of size $N\times M$ i.e., $\phi(x)\in \mathbb{R}^{D\times N\times M}$. This can be interpreted as mapping each pixel at $(n, m)$ to a $D$-dimensional feature embedding $\phi(x)_{nm}$ in the semantic embedding space where knowledge transfer can be facilitated via  word embeddings. The latter computes the inner product between the pixel embedding and word embeddings followed by the softmax that outputs the posterior probability over training classes, %
\begin{align}
\label{eq:prob_spnet}
    P(\hat{y}_{nm} = c|x; W^s) = \frac{\exp (w_c^T \phi(x)_{nm})}{\sum_{c^{\prime}\in \mathcal{S}} \exp (w_{c^{\prime}}^T \phi(x)_{nm})}
\end{align}
where $w_c\in \mathbb{R}^D$ is the $c$-th row of the matrix $W^s$, corresponding to the word embedding of class $c$. For a particular training example $(x, y)$, the standard cross-entropy loss is, 
\begin{align}
\label{eq:loss_spnet}
    \mathcal{L}_{CE} = \sum_{n,m=1}^{N, M}  - \mathds{1}[y_{nm}\neq 0]\log P(\hat{y}_{nm} = y_{nm}|x)
\end{align}
where $y_{nm}$ denotes the true class label at pixel $(n, m)$ and $\mathds{1}[y_{nm}\neq 0]$ is an indicator function that is 1 if $y_{nm}\neq 0$ otherwise 0. Note that the image $x$ might include pixels from unseen classes, but those pixels are not labeled~(i.e., $y_{nm}=0$) and their losses are ignored for ZLSS. The network can be trained in an end-to-end manner by optimizing the above loss on the whole training set $\mathcal{T}$ of seen classes. 

\myparagraph{Inference.} At test time, we predict all classes by searching for the class that yields the highest probability using the word embeddings of seen and unseen classes, 
\begin{align}
    \argmax_{c\in \mathcal{S}\cup \mathcal{U} } P(\hat{y}_{nm} = c|x; [W^s, W^u])
\end{align}

\begin{algorithm}[t]
  $P_t \gets \text{ZLSS model at iteration } t$\;
  $P_{t-1} \gets \text{ZLSS model at previous iteration (t-1)}$\;
  $\{A_1(\cdot),...,A_k(\cdot)\} \gets \text{data augmentations}$\;
  $\mathcal{T} \gets \text{train set}$\;
  $T \gets \text{number of iterations}$\;
  \For{$t=1,2,...,T$}{
  \ForEach{$(x,y)$ in $\mathcal{T}$}{
      $\hat{y} \gets \text{model prediction} P_{t-1}(x)$\;
      $ A \gets \text{augmentations \{}A_1(x),...,A_k(x)\text{\}}$\;
      $ \Gamma \gets \text{hard pseudo labeled masks} \{\bar{y}^k,...\bar{y}^K\}$\;
      $\bar{y} \gets A_1^{-1}(\bar{y}^k) \cap \ldots \cap A_K^{-1}(\bar{y}^K)$\;
      $ \mathcal{L} \gets \mathcal{L}_{CE}(x, y) + \lambda \mathcal{L}_{CE}(x, \bar{y})$\;
      $P_t \gets \text{SGD model update}$\;
      
     }
     $P_{t-1} \gets P_t $
     }
     
  \caption{STRICT pseudo-code}
  \label{pseudocode}
 \end{algorithm}

\subsection{Iterative self-training pipeline}


Figure~\ref{fig:pipeline} shows an overview of our self-training pipeline which consists of two major steps: (1) train the SPNet, and predict pseudo-labels for unlabeled pixels and (2) feed the pseudo-labels back to the training set and retrain the SPNet. The last two steps are iterative, which means that the refined model will be used to generate more accurate pseudo-labels for retraining the model. 
Algorithm \ref{pseudocode} describes our iterative training pipeline in details. 

\myparagraph{Pseudo-label generation.} Given the original training set $\mathcal{T}$ and word embeddings $W^s$ from seen classes, we train the SPNet~\cite{spnet} by optimizing $\mathcal{L}_{CE}$ defined in Equation~\ref{eq:loss_spnet}. Note that the  training set $\mathcal{T}$ contains labeled pixels of seen classes and unlabeled pixels of unseen classes. The unlabeled pixels are ignored when computing  $\mathcal{L}_{SP}$~\cite{spnet}. In other words, unseen classes are actually observed by the network but do not contribute to the training loss, which is a major difference from ZSL. We denote the model learned from this step with $P_0(\hat{y}|x; W^u)$ (see Equation~\ref{eq:prob_spnet}) that outputs the probability distribution over unseen classes. 
Without loss of generality, we assume that the current iteration is t; in this step, the learned model $P_{t-1}$ from the previous iteration $t-1$ is applied to unlabeled pixels to generate pseudo-labeled examples of unseen classes to finetune the model. Here, the unlabeled pixels refer to the pixels in the original training set $\mathcal{T}$ with $y_{nm}=0$. However, pseudo-labels produced by simply making predictions with $P_t$ contain a large amount of label noise which may hurt the model training. To this end, we propose a pseudo-label generator~(denoted $G$) with consistency constraints to filter out potentially wrong pseudo-labels. It is worth noting that the pseudo-label generator makes predictions among only unseen classes~(i.e., zero-label semantic segmentation setting) because it is known that the unlabeled pixels in the training set all belong to unseen classes. Specifically, for each of the training images containing unlabeled pixels, our pseudo-label generator computes $\bar{y}=G(x)$ where $\bar{y}\in \{\{0\}, U\}^{N\times M}$ denotes the pseudo-label mask of unseen classes for the image $x$ and $\bar{y}_{nm}=0$ if the pixel $(n, m)$ belongs to a seen class. The technique details of our pseudo-label generator will be discussed in Section~\ref{sec:consistency}. 

\myparagraph{Iterative self-training.} After the pseudo-label generation step, for each training image $x$, we will have a real label mask $y\in \{\{0\}, S\}^{N\times M}$ of seen classes and a pseudo-label mask $\bar{y}\in \{\{0\}, U\}^{N\times M}$ of unseen classes. $y$ and $\bar{y}$ are then used to compute the labeled and pseudo-labeled loss terms separately. Formally, we optimize the following combined loss to finetune the model $P_{t-1}$, 
\begin{align}
    \mathcal{L} = \mathcal{L}_{CE}(x, y) + \lambda \mathcal{L}_{CE}(x, \bar{y})
\end{align}
where $\mathcal{L}_{CE}$ is the cross-entropy loss defined in Equation.~\ref{eq:loss_spnet} and $\lambda$ is a hyperparameter. The first loss term is the same with the original SPNet loss while the second term is computed on unlabeled pixels with the generated pseudo-labels being the supervision. The main insight is that the pseudo-labeled pixels circumvent the need of unseen class data, yielding a more balanced training set that facilitates the GZLSS. Intuitively, the better the model is, the more accurate pseudo-labels it can generate. Therefore, we can apply the finetuned model $P_t$ from current iteration $t$ back to generate pseudo-labels for the next iteration $t+1$. The finetuning and pseudo-label generation steps run iteratively  according to the procedure described in Section \ref{sec:validation}.






\subsection{Consistency constraints}
\label{sec:consistency}

Generating hard pseudo-labels~(one-hot prediction) directly from the model $P_t$ is not ideal because the noise level of one single prediction is high~(as shown in Figure~\ref{fig:teaser}). CagNet~\cite{cagnet} and ZS5~\cite{zs3} address this issue by filtering out the pixels for which no unseen classes are present among the top p\% softmax activations. Therefore, even if they ignore the predictions seen for the unlabelled pixels they obtain the softmax activations for them in the GZS scenario, hence including seen classes to the search space of unlabelled pixels that for sure are background or unseen. Moreover, their strength relies on the hyperparameter $p$, to be finetuned according to the confidence of the pseudo-label generator and to the dataset. 
We propose a simple approach to reduce the noise of pseudo-labels based on the consistency regularization~\cite{crossConsistency,cons1,cons2,cons3,cons4}. The key assumption is that a pseudo-label is more likely to be correct if the model makes the same prediction from multiple augmented variants of the image. More formally, given an image $x$ containing unlabeled pixels, we apply $K$ different data augmentations~(denoted as $A_k(\cdot)$) to obtain a set of $K$ augmented images i.e., $\{A_1(x), \ldots, A_K(x)\}$. Here we denote $A_1$ as an identity mapping and mainly consider the  horizontal mirroring and scaling with different scaling factors as our data augmentation scheme since we find they work best in the experiments. We then apply the model $P_t$ to generate hard pseudo-labels~(one-hot labels) for every unlabeled pixels in each augmented image, 
\begin{align}
   \bar{y}^k_{nm} &= \argmax_{c\in \mathcal{U} } P(\hat{y}_{nm} = c|A_k(x); W^u) \\ \nonumber
   &\forall k\in \{1,\ldots, K\}, \forall (n,m) \in \mathcal{I}.
\end{align}
This yields a set of $K$ hard pseudo-labeled masks $\{\bar{y}^k, \ldots,  \bar{y}^K\}$ for the image $x$. Intuitively, those data augmentations only transform the image spatially and the semantic of each pixel should remain the same. We then obtain the final pseudo-label mask by applying the intersection operation on those $K$ masks,
 \begin{align}
   \bar{y} =  A_1^{-1}(\bar{y}^k) \cap \ldots \cap A_K^{-1}(\bar{y}^K),
\end{align}
where $A_k^{-1}$ denotes the inverse data augmentation that transforms augmented masks back to the original coordinates. The intersection operation essentially filters out the pseudo-labels that are inconsistent across multiple augmented masks, yielding more accurate pseudo-labels. Although a similar consistency regularization has been explored in semi-supervised learning~\cite{berthelot2019mixmatch}, we are the first to apply the consistency constraints for the GZLSS task.

\section{Experiments}
\label{sec:validation}

\myparagraph{Datasets and metrics.} We evaluate our approach on two datasets, PascalVOC12 \cite{pascalvoc12} and COCO-stuff \cite{cocostuff}, following previous works~\cite{spnet,cagnet} for the data splits and the validation procedure. {PascalVOC12} is an object segmentation benchmark, containing images of 20 foreground objects plus the background class. 
{COCO-stuff} is a large-scale dataset for scene segmentation, with 164K images containing 80 common objects and 91 stuff classes. Our train/val/test sets are mutually exclusive classes i.e., 11185/500/1449 images from 12/3/5 classes on VOC12 and 116287/2000/5000 images from 155/12/15 classes on COCO-stuff. As train and val sets belong to disjoint subsets of seen classes, we use the following two stage procedure for fine tuning: (i) we first select the best hyperparameters considering as seen the train classes and as unseen the validation ones; (ii) we perform training considering as seen both train and validation classes with \textit{fixed} hyperparameters (i.e. without looking at the validation set again).
For PascalVOC12, we perform additional experiments where the background is included among the set of seen classes. 
 Following \cite{spnet}, we measure the generalized zero-label performance in terms of mean Intersection over Union (mIoU) on the seen (S) and unseen (U) classes, as well as the harmonic mean (HM) among the two. 

\myparagraph{Baselines and implementation details.}
We compare our approach with three state-of-the-art GZLSS methods, namely the baseline SPNet \cite{spnet}, and two generative approaches, ZS3 \cite{zs3} and CaGNet \cite{cagnet}. Additionally, we  include the self-training variants of CaGNet (CaGNet+ST) and ZS3 (ZS5), both using the top percentage of the pixels assigned to unseen classes as pseudo-labels. Moreover, we report the results of another baseline, the calibrated SPNet trained by performing hard pseudo-labelling on unlabeled pixels, without any consistency strategy (SPNet+ST). For fair comparison with previous works, we use DeepLabV2 \cite{deeplab1} as the segmentation model with an Imagenet \cite{imagenet} pretrained ResNet-101 \cite{resnet} as backbone. We train our network with SGD, with a momentum of 0.9 and a weight decay of $5\cdot10^{-4}$. The learning rate is initially set to $2.5 \cdot 10^{-4}$ with a polynomial decay, as in \cite{deeplab1}. After training the base model  train the model for 2K iterations on PascalVOC12 and for 22K iterations for COCO-stuff, using a batch-size of 8 images. After training the network for 20K iterations for VOC and 100K on COCO with only supervision on seen class pixels, we keep the same hyperparameters and we fine-tune the network with our self-training strategy, considering one cycle of self-training finished after 2K iterations on PascalVOC12 and after 22K iterations for COCO-stuff. Results for SPNet and SPNet+ST are reported after running the approach under our framework. 

{
\setlength{\tabcolsep}{3.5pt}
\renewcommand{\arraystretch}{1.2}
\begin{table}[t]
\centering
\resizebox{\columnwidth}{!}{%
\begin{tabular}{l|c c c|c c c}
    \multirow{2}{*}{Method}& \multicolumn{3}{c|}{\textbf{PascalVOC12}} & \multicolumn{3}{c}{\textbf{COCO-stuff}}\\

 & S & U & HM & S & U & HM \\
\hline
SPNet \cite{spnet} & 73.3 & 15.0 & 21.8 & 20.5 & 14.3 & 16.8\\
ZS3 \cite{zs3} & 77.3 & 17.7 & 28.7 & 34.7 & 9.5 & 15.0 \\
CaGNet \cite{cagnet} & 78.4 & 25.6 & 39.7 & 35.5 & 12.2 & 18.2\\
\hdashline
{SPNet+ST} & 77.8 & 25.8 & 38.8 & 34.6 & 26.9 & 30.3\\
ZS5 \cite{zs3} & 78.0 & 21.2 & 33.3 & 34.9 & 10.6 & 16.2\\
CaGNet + ST \cite{cagnet} & 78.6 & 30.3 & 43.7 & \textbf{35.6} & 13.4 & 19.5\\\hline
\textbf{STRICT} & \textbf{82.7} & \textbf{35.6} & \textbf{49.8} & 35.3 &	\textbf{30.3} & \textbf{32.6}
\end{tabular}}
\caption{Comparing with the state of the art on PascalVOC12 and COCO-stuff.}
\label{tab:sota}
\end{table}
}

\subsection{Comparison with the state of the art}

We compare our approach with the state of the art on both PascalVOC12 and COCO-stuff, reporting the results in Table \ref{tab:sota}. A first outcome of the experiments is that self-training strategies improve the performance of all methods and for all metrics. In particular, in PascalVOC12, ZS5 improves his not self-trained counterpart by almost 5\% in HM and CaGNet+ST improves CaGNet by almost 4\% on the same metric. Remarkably, SPNet+ST improves the base SPNet by 17\% in HM, with a 11\% improvement on the unseen classes and a 4.5\% on the seen ones, on PascalVOC12. Similar observations hold for COCO-stuff, where the improvements on the HM are of 1.5\% for ZS5 over ZS3, 1.3\% of CaGNet over CaGNet and 13.5\% of SPNet+ST over SPNet.  Note that, with this simple strategy, SPNet surpasses all more complex generative approaches on COCO-stuff for unseen mIoU and HM, while achieving a lower harmonic mean than CaGNet on PascalVOC12. These results confirm that considering the co-occurrence of seen and unseen classes through self-training is always very beneficial. Moreover, the improvements are larger in non-generative methods (SPNet) than in generative ones. 
A second, clear outcome, is that our STRICT strategy outperforms every published results by a good margin. On PascalVOC12, it surpasses of 6.1\% on harmonic mean and of 5.3\% on unseen mIoU the previous state of the art (CaGNet+ST). The margin is even higher in the large scale COCO-stuff dataset, with our approach surpassing CaGNet of 16.9\% on unseen class mIoU and of almost 13.1\% on harmonic mean. If we compare STRICT with the SPNet+ST baseline, we see that the improvement is higher on PascalVOC12 (4.9\% on seen classes, 9.8\% on unseen classes and 11\% on harmonic mean) while being less marked on COCO-stuff, with a 0.7\% improvement on seen class, 3.4\% improvement on unseen class mIoU and 2.3\% on the harmonic mean. 
These improvements are outstanding, confirming the importance of equipping any zero-label semantic segmentation model with an effective self-training strategy. Note that the self-training approach greatly reduces the bias of the network on seen classes. Indeed, differently form SPNet, we do not need a calibration term to balance seen and unseen class predictions. Similarly, different from generative approaches, we do not rely on synthesised pixels, but rather exploit the more precise information coming from the unlabeled pixels of our images.  


{
\setlength{\tabcolsep}{5pt}
\renewcommand{\arraystretch}{1.2}
\begin{table}[t]
\centering
\resizebox{0.65\columnwidth}{!}{%
\begin{tabular}{l|c c c}
  \multirow{2}{*}{Method}  & \multicolumn{3}{c}{\textbf{PascalVOC12}} \\
 & S & U & HM \\
\hline
SPNet \cite{spnet} & 54.7 & 2.5 & 4.7 \\
ZS3 \cite{zs3} & 59.0 & 4.0 & 7.5 \\\hdashline
{SPNet+ST} & 72.7&4.0 & 7.6\\
ZS5 & 66.1 & 1.7 & 3.7\\
\hline
\textbf{STRICT}  & \textbf{74.7} & \textbf{14.3} & \textbf{24.0}
\end{tabular}
}
\caption{PascalVOC12 results with background class included among the seen set.}
\label{tab:background}
 \vspace{-8pt}
\end{table}
}

{
\setlength{\tabcolsep}{5pt}
\renewcommand{\arraystretch}{1.2}
\begin{table}[t]
\centering
\resizebox{0.8\columnwidth}{!}{%
\begin{tabular}{ c c | c c c}
 Mirroring & Scaling & S & U & HM  \\
\hline
& & 77.8 & 25.8 & 38.8 \\
\hdashline
\cmark &  & 80.4 & 27.2 & 40.7 \\
\hdashline
&down & 82.1 & 27.8 & 41.5 \\
&up & 82.0 & 31.1 & 45.1 \\
&random & 81.6 & 29.4 & 43.2\\
\hdashline
\cmark &down & \textbf{83.7} & 29.2 & 43.3 \\
\cmark &up & 82.5 & \textbf{32.9} & \textbf{47.0} \\
\cmark &random & 83.2 & 31.4 & 45.6
\end{tabular}}
\caption{Ablation of different transformations for the consistency constraint of STRICT on PascalVOC12.}
\label{tab:perturbations}
\end{table}
}

\begin{figure*}
     \centering
     \begin{subfigure}[t]{0.48\textwidth}
         \centering
    \includegraphics[width=\linewidth]{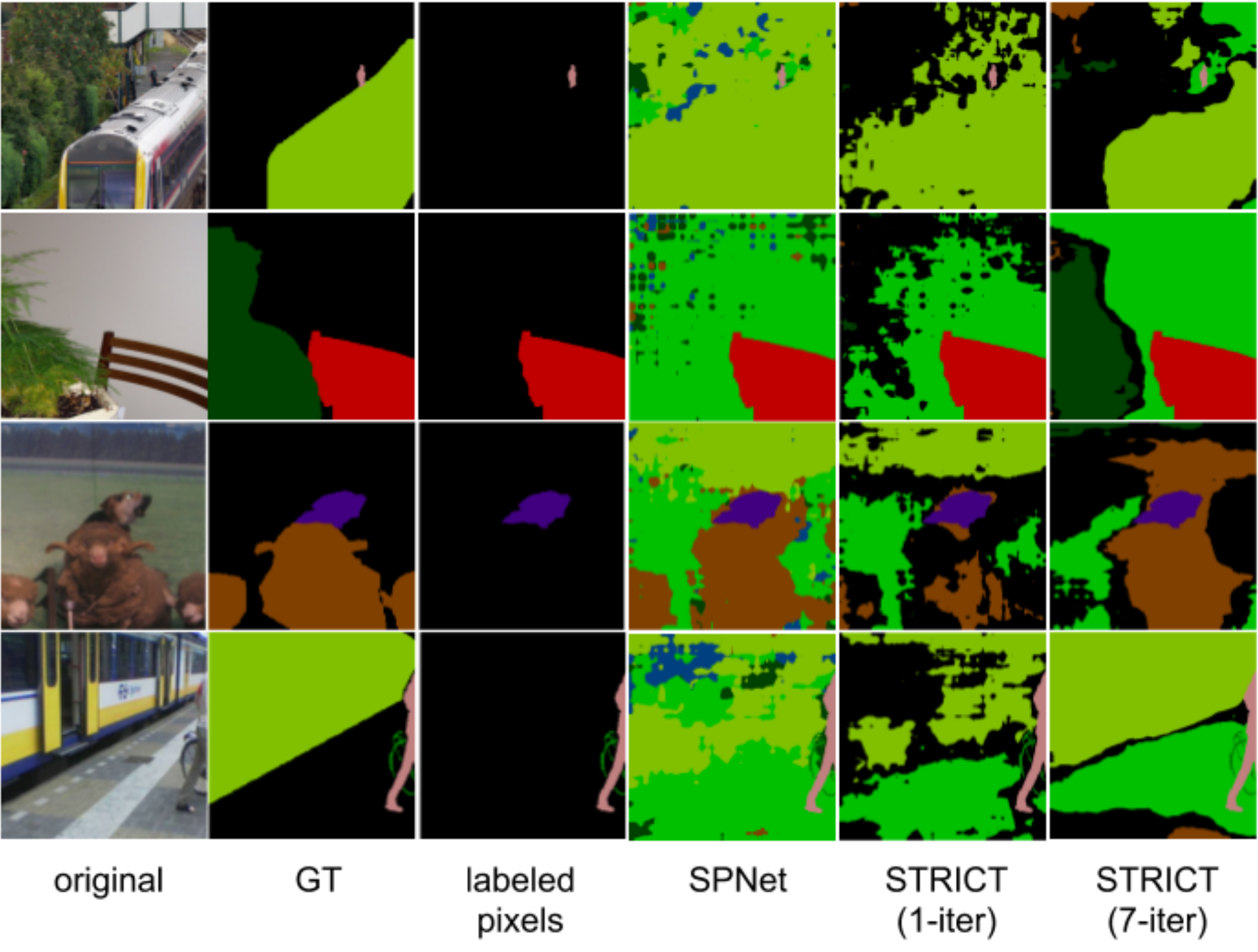}
\caption{Pseudo-labels generated with STRICT  for PascalVOC12 unseen classes when background is ignored.}
\label{fig:pseudoNOBKG}
     \end{subfigure}
     \hfill
     \begin{subfigure}[t]{0.48\textwidth}
         \centering
  \includegraphics[width=\linewidth]{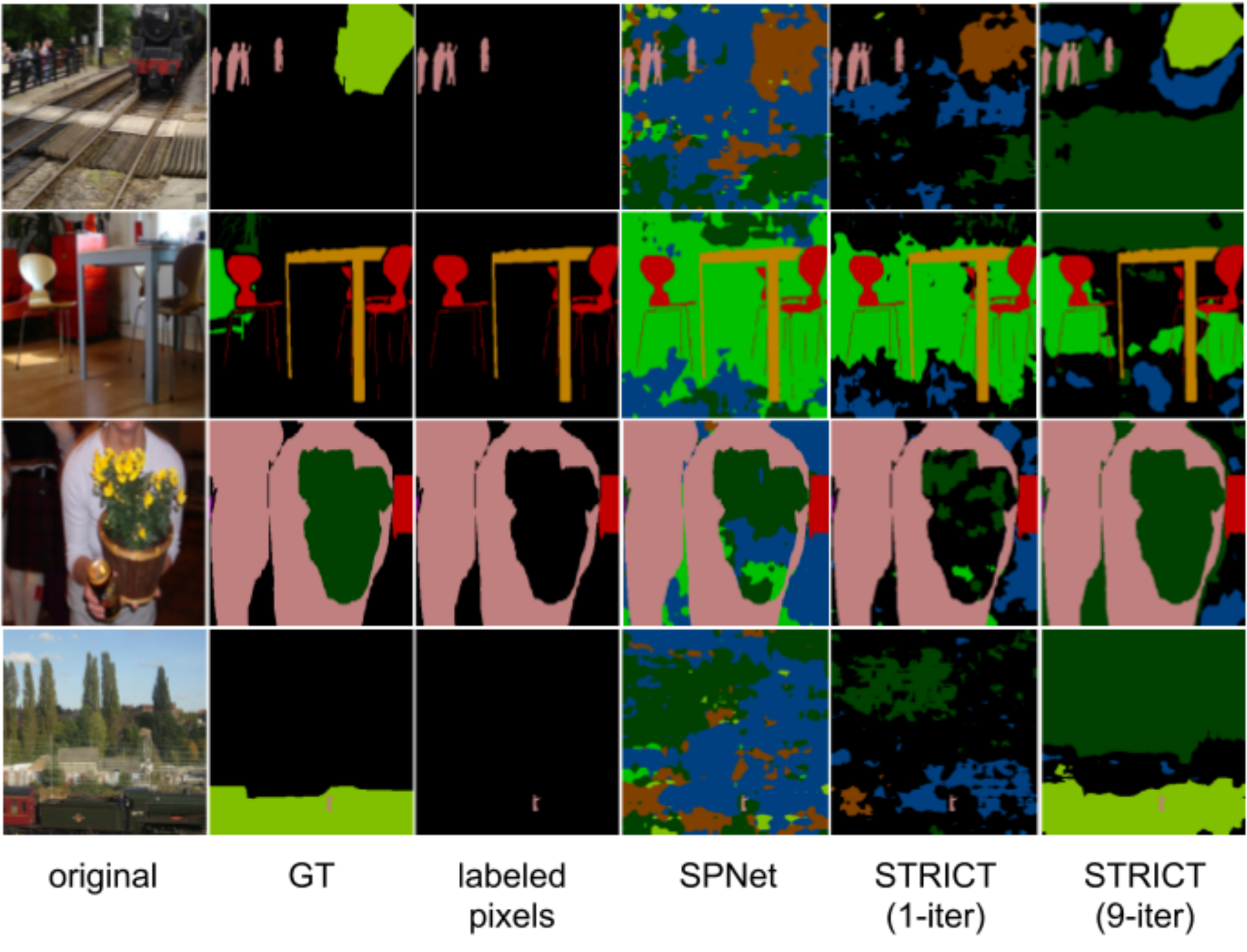}
\caption{Pseudo-labels generated with STRICT for PascalVOC12 unseen classes when background is included. }
     \label{fig:pseudoBKG}
     \end{subfigure}
     \caption{Qualitative pseudo-labeling results of STRICT on PascalVOC12 without (left) and with (right) background as seen class. Train GT refers to labels for the unseen classes.}
\end{figure*}

\myparagraph{Impact of the background on PascalVOC12.}
Standard GZLSS approaches for object segmentation usually do not consider the distinction between foreground and background, performing evaluation only on pixels of foreground objects. Here we evaluate the change in performance when the background class is included in the search space. 
Note that this scenario is far more challenging since the pixels of unseen classes might be labeled as background, hampering the model capability to discriminate them. 

Results are reported in Table \ref{tab:background}, for our method, SPNet, ZS3 and their self-trained variants. All methods, both STRICT and the baselines, suffer a severe performance degradation when including the background in the classifier. Indeed, if we compare Table \ref{tab:sota} with Table \ref{tab:background}, we can see how SPNet achieves only 2.5\% of mIoU on unseen classes (almost 12\% lower than Table \ref{tab:sota}) with an overall 4.7\% on harmonic mean (17\% lower). With self-training, results improve only slightly, with SPNet+ST obtaining 4\% mIoU on unseen classes and a 7.6\% of harmonic mean. Surprisingly, ZS3 outperforms its self-trained counterpart ZS5 in this setting as learning a robust classifier for unseen classes in a generative fashion is difficult in segmentation, due to the high complexity of the images. Additionally, both the generation and the pseudo-labeling process are hampered by the bias of the network toward predicting background in place of unseen class pixels. 
Our STRICT approach 
is effective even in this setting, with an mIoU on unseen classes of 14.3\% and an overall harmonic mean of 24\%. We highlight how these results are on par with the performance of the calibrated SPNet on the standard scenario where the background is ignored, being slightly lower (3\% on harmonic mean and unseen class mIoU) than ZS3. Despite these promising results, the gap in performance among our model on the two scenarios is still large (25\% on harmonic mean and 21\% on unseen mIoU). This means that the technical challenges of GZLSS for object segmentation when background is included require additional technical components, explicitly addressing problems such as the semantic shift of the background class \cite{cermelli2020modeling}.


\begin{figure}[t]
    \centering
        \includegraphics[width=1\linewidth]{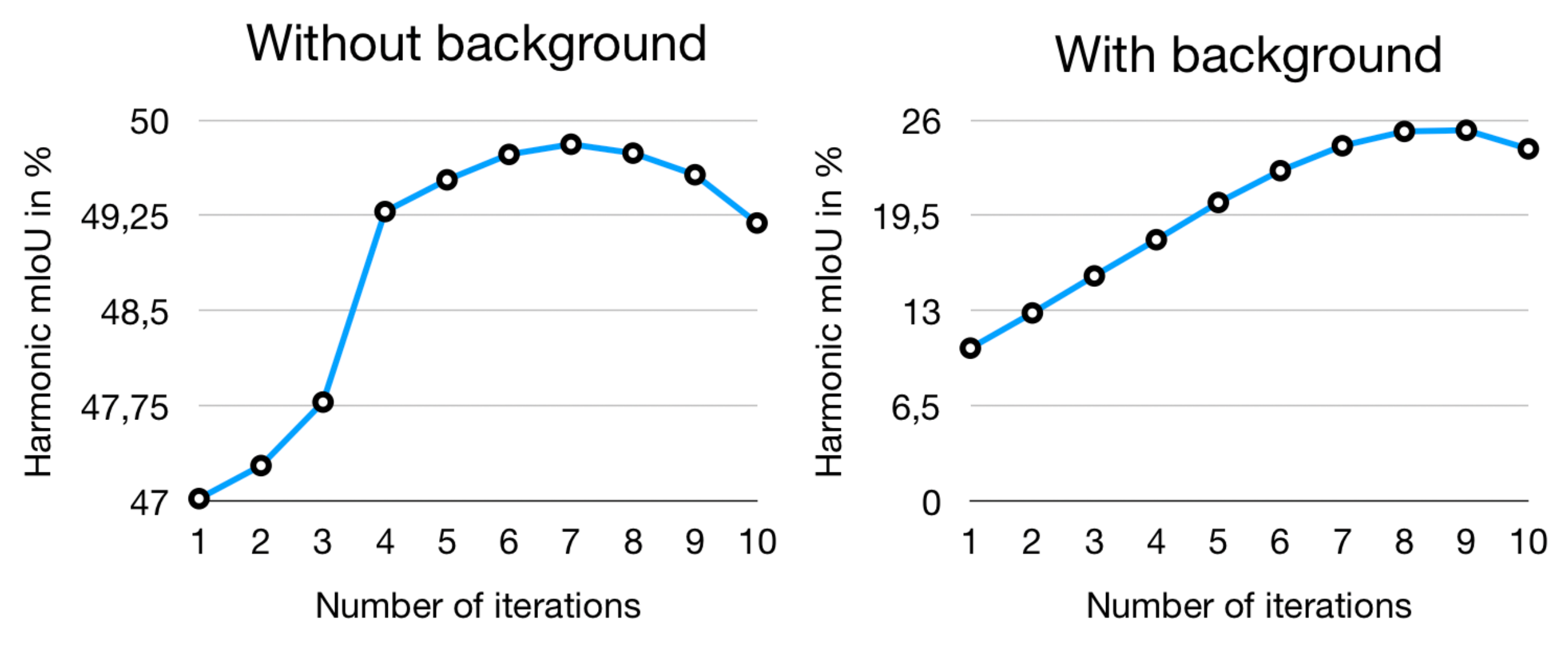}
    \caption{STRICT mIoU along with the number of iterative fine-tuning steps $i$.}
\label{fig:iterations}
\end{figure}

\subsection{Ablation study} 
\label{ablation}

In this section, we study the effect of different components of our approach, namely the type of transformations applied for the consistency procedure and the number of self-training iterations on PascalVOC12.

\myparagraph{Different image transformations.}
We first evaluate which image transformations are more effective for applying our self-training with consistency constraints. In particular, we consider simple and invertible image-level transformation such as three variants of multi-scaling (down, up and random scaling) and mirroring. 
We report the results of our analysis in Table \ref{tab:perturbations}.
As the table shows, performing multi-scaling is, in general, more effective than applying only mirroring. Among the scaling alternatives, upscaling brings the best results, with the highest mIoU on unseen classes (31.1\%) and harmonic mean (45.1\%). Combining mirroring and upscaling, we obtain the best performance, with 32.9\% of mIoU on unseen classes and 47\% of harmonic mean.


\myparagraph{Number of self-training iterations.} An important aspect of our algorithm is the iterative self-training procedure, with the pseudo-labeling model updated after each iteration. Here, we analyze the impact of the number of self-training iterations for STRICT in Figure \ref{fig:iterations}, where we report the results as mIoU on unseen classes and harmonic mean and in both cases with and without background included. As the Figure shows, for both metrics and settings performances tend to increase as the number of self-training iterations does. In particular, performances rapidly increase until six self-training iterations, after which they saturate and/or slightly decrease. This decrease can be caused by the fact that we do not have ground-truth for the unseen class pixels and noisy predictions can be reduced but not entirely eliminated by our consistency constraint. 

\begin{figure*}
\begin{center}
  \includegraphics[width=\linewidth]{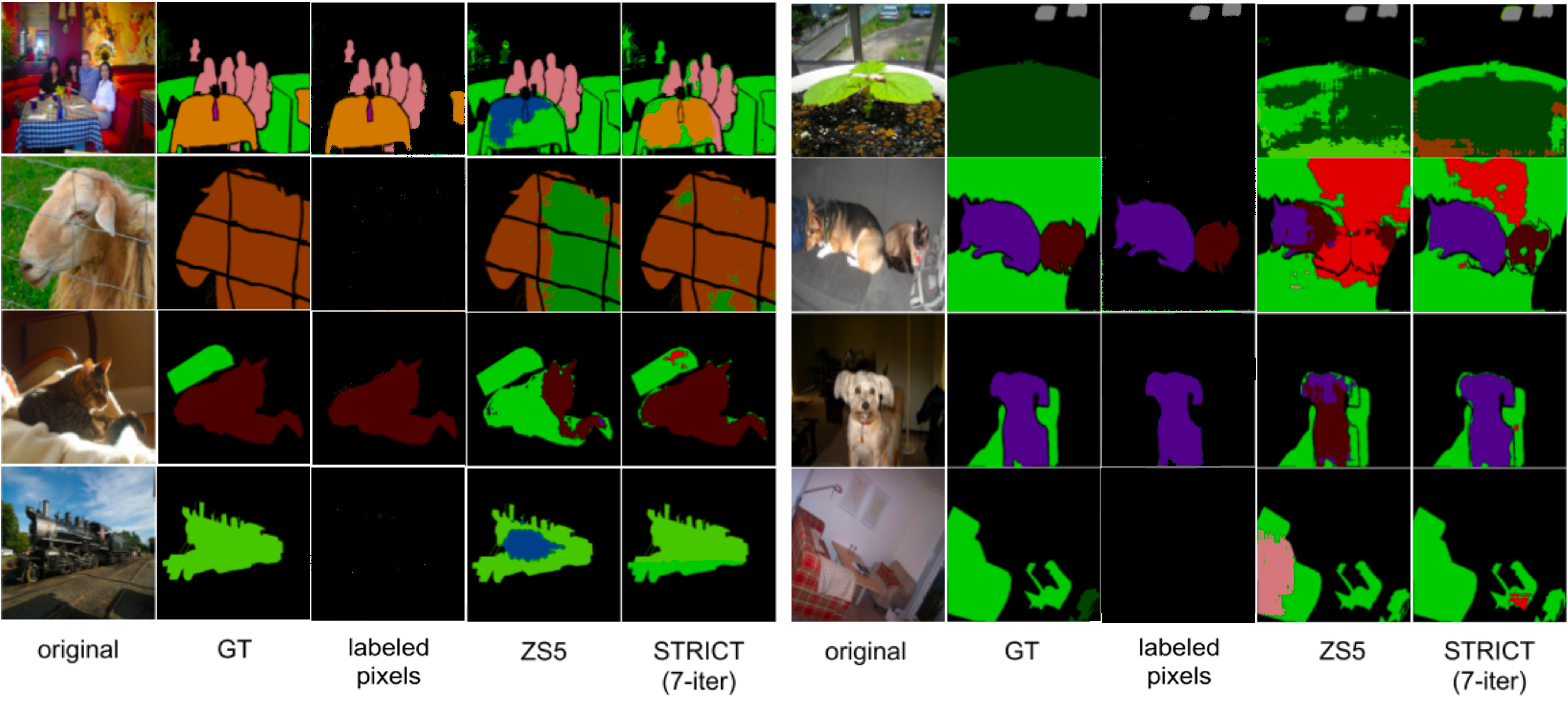}
\end{center}
\vspace{-20pt}
  \caption{Qualitative results of STRICT with the original SPNet and ZS5 on PascalVOC12 when the background is ignored. }
\label{fig:finalNOBKG}
\vspace{-10pt}
\end{figure*}

\subsection{Qualitative analysis}
In this section, we report qualitative analysis on PascalVOC12 regarding i) the pseudo-labels generated by our model and ii) semantic segmentation results.

\myparagraph{Pseudo-labels.} Another crucial point of our algorithm is generating good pseudo-labels as supervision signal for our model on unseen class pixels. Figure \ref{fig:pseudoNOBKG} and \ref{fig:pseudoBKG} show some annotations on unseen classes obtained by our model, when the background is ignored and included during training respectively. For each original image, \textit{GT} is the actual ground truth $y$, while \textit{labeled pixels} represent the annotation for seen classes $y^s$ that the model sees before the pseudo-labeling. From the Figures we see that, while our starting point (SPNet+ST) detects the presence of pixels of unseen classes, the predictions are noisy, with some pixels assigned to classes not present in the current image. Our consistency constraint (STRICT) allows to largely reduce the noise, eliminating most of the pseudo-labels assigned to pixels of classes not present in the current image (\eg
\textit{train} in third row of Figure \ref{fig:pseudoNOBKG}, \textit{tv/monitor} in first and fourth rows of Figure \ref{fig:pseudoBKG}). With more iterations, STRICT produces more refined pseudo-labels, where spatially coherent structure are present. This means the pseudo-label generator capture global information of unseen classes, something which is not possible to do with a single stage of pseudo-labeling. 

\myparagraph{Semantic segmentation.}
Finally, to compare our model with the other baselines, we show qualitative semantic segmentation results of our method and ZS5 on Figure \ref{fig:finalNOBKG}. As the Figure shows, our model is able to correctly identify pixels of unseen (\eg \textit{sofa}) as well as seen (\eg \textit{person}) categories. Moreover, it achieves a good trade-off between seen and unseen classes. For instance, on the image of the second row, left, ZS5 misclassifies most of the pixels of the \textit{sheep} (unseen class) as a \textit{cow} (seen class), showing its bias toward seen classes. On the other hand, our model segments almost perfectly the \textit{sheep}, with few pixels misclassified. A similar example is the \textit{table} (unseen class) misclassified by ZS5 as \textit{tv/monitor} while almost correctly segmented by our model. These images show also some drawbacks of our approach, meaning the dependency of the results on the number of co-occurring pixels. For instance, since \textit{plant} occupies low portions of the images, it is hard for the network to produce consistent pseudo-labels for it, with consequently low recognition ability of the final model for that class. Future works might exploit strategies to regularize the supervision for unseen classes based on the number of pseudo-labels generated for each of them.

\section{Conclusions}
In this work, we proposed a self-training approach to learn the model to segment classes not annotated in the training set by leveraging on their semantic representation.
Our self-training pipeline is simple, robust and highly scalable, as it relies on the ability of the model to predict consistent labeling among different augmented versions of the same image to filter the generated pseudo-labels and on the iterative strengthening of the pseudo-label generator.
We demonstrated the effectiveness of this method on two commonly used benchmarks for semantic segmentation and we obtained that applying such simple considerations outperforms other more complex strategies in the GZLSS.

\section*{Acknowledgments}
This work has been partially funded by the ERC 853489 - DEXIM, the ERC grant N. 637076 RoboExNovo, and by the DFG – EXC number 2064/1 – Project number 390727645.

{\small
\bibliographystyle{ieee_fullname}
\bibliography{egbib}
}

\end{document}